\documentclass[conference]{IEEEtran}
\IEEEoverridecommandlockouts
\usepackage{amsmath,amssymb,amsfonts}
\usepackage{algorithmic}
\usepackage{graphicx}
\usepackage{textcomp}
\usepackage{xcolor}
\def\BibTeX{{\rm B\kern-.05em{\sc i\kern-.025em b}\kern-.08em
    T\kern-.1667em\lower.7ex\hbox{E}\kern-.125emX}}
\usepackage{natbib}
\setcitestyle{numbers,square}
\usepackage{algorithm}  
\usepackage{algorithmic}
\usepackage[utf8]{inputenc}
\usepackage{soul}
\usepackage{booktabs}

\usepackage{amsmath}
\usepackage{amsthm}
\usepackage{amssymb}
\usepackage{graphicx}

\usepackage{color}
\usepackage{xcolor}
\usepackage{subfigure}
\usepackage{bm}

\usepackage{colortbl}
\definecolor{mygray}{gray}{.8}
\definecolor{mypink}{rgb}{.99,.91,.95}
\definecolor{mycyan}{cmyk}{.3,0,0,0}

\def\ie{\textit{i.e.}}
\usepackage[colorlinks,linkcolor=black,citecolor=black]{hyperref}

\usepackage{algorithm}
\usepackage{algorithmic}

\graphicspath{{figures/}}
\usepackage{multirow}

\usepackage{newfloat}
\usepackage{listings}
\floatstyle{ruled}
\newfloat{listing}{tb}{lst}{}
\floatname{listing}{Listing}

\begin{document}

	\title{Constructing Sample-to-Class Graph for Few-Shot
Class-Incremental Learning}

\author{
	\IEEEauthorblockN{
		Fuyuan Hu$^{1}$, 
		Jian Zhang$^{1}$, 
		Fan Lyu$^{2*}$, 
		Linyan Li$^{3}$,
		Fenglei Xu$^{1}$} 
	\IEEEauthorblockA{$^{1}$ Suzhou University of Science and Technology, $^{2}$ Tianjin University, $^{3}$ Suzhou Institute of Trade \& Commerce, 
	\\  \{fuyuanhu@mail, jianzhang@post, xufl@mail\}.usts.edu.cn, lilinyan@szjm.edu.cn, fanlyu@tju.edu.cn}
}


\maketitle

\begin{abstract}
  Few-shot class-incremental learning (FSCIL) aims to build machine learning model that can continually learn new concepts from a few data samples, without forgetting knowledge of old classes. 
  The challenges of FSCIL lies in the limited data of new classes, which not only lead to significant overfitting issues but also exacerbates the notorious catastrophic forgetting problems. As proved in early studies, building sample relationships is beneficial for learning from few-shot samples. In this paper, we promote the idea to the incremental scenario, and propose a Sample-to-Class (S2C) graph learning method for FSCIL. 
  Specifically, we propose a Sample-level Graph Network (SGN) that focuses on analyzing sample relationships within a single session. This network helps aggregate similar samples, ultimately leading to the extraction of more refined class-level features. 
  Then, we present a Class-level Graph Network (CGN) that establishes connections across class-level features of both new and old classes. This network plays a crucial role in linking the knowledge between different sessions and helps improve overall learning in the FSCIL scenario. Moreover, we design a multi-stage strategy for training S2C model, which mitigates the training challenges posed by limited data in the incremental process. 
  The multi-stage training strategy is designed to build S2C graph from base to few-shot stages, and improve the capacity via an extra pseudo-incremental stage.
    Experiments on three popular benchmark datasets show that our method clearly outperforms the baselines and sets new state-of-the-art results in FSCIL. The code is available at \href{https://github.com/DemonJianZ/S2C}{github.com/DemonJianZ/S2C}.
\end{abstract}






\maketitle

\renewcommand{\thefootnote}{\fnsymbol{footnote}}
\footnotetext[1]{Corresponding author.}

\section{Introduction}
\begin{figure}[ht]
	\centering
	\includegraphics[width=1.\linewidth]{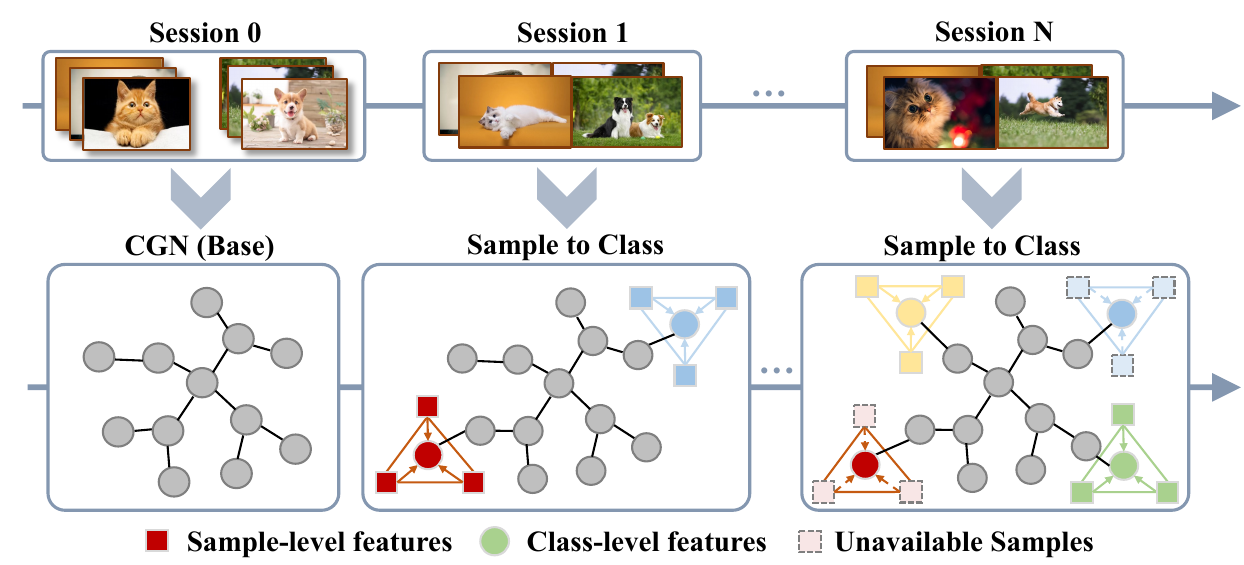}
	\caption{Illustration of our proposed S2C for FSCIL. 
		\textbf{Top}: the setting of FSCIL.	
		\textbf{Bottom}: Sample-level to Class-level graphs.
	}
	\label{fig:motivation}	
\end{figure}

The volume of data on the internet is constantly increasing, and in response to this growing data, incremental learning~\cite{zhou2021learning} has seen significant development in recent years. 
When new data is labeled for new classes, it introduces the challenge of Class-Incremental Learning (CIL)~\cite{pham2021dualnet,zhao2022deep,wang2022learning}, and a prominent issue that emerges is catastrophic forgetting~\cite{rebuffi2017icarl}. 
The catastrophic forgetting refers to the decline in discriminative ability for previously learned classes. 
While many solutions to CIL involve abundant training samples~\cite{gomes2017survey}, practical applications sometimes have only few samples, because of the challenges of data collection or labeling.
For example, in scenarios involving personalized content recommendations while considering user privacy, the available data is often severely limited. 
This scenario of CIL with few training samples is termed Few-Shot Class-Incremental Learning (FSCIL)~\cite{Topology}. 
Similar to CIL, learning new classes in FSCIL results in catastrophic forgetting of prior classes. Furthermore, due to the scarcity of instances from new classes, \emph{overfitting} tends to occur on these restricted inputs. This, in turn, heightens the learning difficulty of incremental tasks.
As shown in Fig.~\ref{fig:motivation}, the training of FSCIL is class-incremental and in sequence, and the data of past classes is unavailable. The incremental model is evaluated across all previously encountered classes at any sessions. 

When addressing FSCIL challenges, one plausible approach is to employ traditional CIL methods, including widely used techniques like knowledge distillation~\cite{CILKD}. 
While CIL approach has partially alleviated the problem of catastrophic forgetting, straightforwardly adopting there methods in FSCIL is ill-advised, given the scarcity of training samples that leads to overfitting and inadequate performance on previously learned classes~\cite{tao2020few}.
On the other hand, for each few-shot session, another approach is to applied Few-Shot Learning (FSL) methods to the current few samples. For example, as proved in~\cite{zhang2021few,zhou2022forward}, using class means (prototype features) to mitigate overfitting is effective in FSL. 
In several recent FSL works~\cite{EGNN}, building sample relationships using Graph Neural Network (GNN)~\cite{GNN} is beneficial for learning from very few samples.
GNN can express complex interactions between samples by performing feature aggregation from neighbors, and mining refined information from a few samples between support and query data. 
However, these FSL methods ignore the incremental sessions, and show unacceptable catastrophic forgetting.
In summary, current FSCIL methods face a challenge in balancing the effective learning of new tasks with the forgetting suppression of old tasks. 
But some of these methods~\cite{cheraghian2021semantic,MetaFSCIL,CILKD} focus on bringing techniques from CIL to suppress catastrophic forgetting, while some others~\cite{pernici2021class,SDC,zhu2021self} aim to enhance model adaptation for few-shot tasks, thus they could hardly effectively address both aspects in FSCIL.

Inspired by the use of GNN in FSL, in this paper, we investigate to build the relationships of cross-session classes using limited samples in FSCIL, aiming to enhance the performance of individual few-shot tasks and reduce the forgetting at the same time.
As shown in Fig.~\ref{fig:motivation}, this paper introduces an innovative \emph{Sample-to-Class (S2C)} graph learning approach, which establishes connections from the sample level to the class level.
\textbf{The model}: The S2C model has two major components to build graph relations from sample-level to class-level. First, the Sample-level Graph Network (SGN) evaluates the similarity between samples within a single few-shot session, clusters samples from the same class, and distinguishes samples from different classes. The SGN yields more refined features and mitigates the overfitting problem to some extent. 
Moreover, to construct the semantic relationship among multiple classes from different sessions during incremental learning, we propose a Class-level Graph Network (CGN). The CGN forges connections between old and novel classes, thereby augmenting the capacity to differentiate classes across sessions and alleviating the catastrophic forgetting. 
\textbf{The training}: To smoothly deploy the S2C model in FSCIL, we propose a novel training strategy, which comprises three main stages. The first stage takes advantage of the ample training data available in the base session to initialize the CGN, thereby preserving a substantial amount of prior knowledge for the subsequent learning of few-shot tasks. The second stage is designed to address the issue of insufficient sample-level relationship mining due to the limited number of samples. This is achieved through the S2C pseudo incremental learning, which adapts the S2C model to the FSL task beforehand. During this pseudo-incremental process, FSL tasks are randomly sampled from the base dataset, and virtual FSCIL tasks are generated. In the last stage, we deploy the S2C model to a real FSCIL scenario for further optimisation.

Our contributions can be summarized in three main aspects:
\begin{enumerate}[1.]
\item We introduce a novel S2C method for FSCIL, comprising the SGN and the CGN. This innovative structure serves to bridge the relationships between old and new classes at two distinct levels. To the best of our knowledge, our work pioneers the incorporation of graph neural networks into FSCIL from two unique perspectives.
\item We propose a novel S2C multi-stage training strategy, which trains the S2C model incrementally, allowing S2C to adapt and construct graphs effectively even with limited samples. With the three stages, S2C establishes semantic relationships across multiple sessions, mitigating the issue of catastrophic forgetting.
\item We conduct comprehensive experiments on benchmark datasets, including CIFAR100, miniImageNet, and CUB200. The empirical results substantiate the superiority of our approach over state-of-the-art methods, demonstrating a substantial performance margin.
\end{enumerate}

\section{Related Work}

\noindent
\textbf{Few-Shot Learning.}
Few-shot learning aims at rapidly generalizing to new tasks with limited samples, 
leveraging the prior knowledge learned from a large-scale base dataset.
The existing methods can be divided into two groups.
Optimization-based methods~\cite{lee2019meta,rusu2018meta,BTVR} try to enable fast model adaptation with few-shot data.
Metric-based algorithms~\cite{zhang2020deepemd,ma2021transductive,MetaSearch} utilize a pretrained backbone for feature extraction, and employ proper distance metrics between support and query instances.
Recent research tries to leverage GNNs to explore complex similarities among examples. 
DPGN~\cite{yang2020dpgn} builds up a dual graph to model distribution-level relations of examples for FSL. 
ECKPN~\cite{chen2021eckpn} proposes an end-to-end transductive GNN to explore the class-level knowledge.

\noindent
\textbf{Meta-learning.}
Meta-learning is commonly described as the concept of "learning to learn." This approach involves the extraction of knowledge and insights from multiple learning episodes and then leveraging this acquired experience to enhance performance in future learning tasks ~\cite{Metalearning}. Meta-learning is typically divided into two distinct stages. In the first stage, known as the meta-training stage, a model is trained using multiple source or training tasks. This training process aims to acquire initial network parameters that exhibit robust generalization capabilities. In the second stage, known as the meta-testing stage, new tasks are introduced, and the conditions for these tasks are identical to those of the source tasks. Meta-learning is inherently well-suited for FSL, and numerous research studies have employed meta-learning as an approach for FSL. This enables models to acquire knowledge and adapt from a limited number of samples associated with new tasks ~\cite{MetaFSCIL, CalibratingCNN}.

\noindent
\textbf{Class-Incremental Learning.}
Class-Incremental Learning aims to learn from a sequence of new classes without forgetting old ones, which is now widely discussed in various computer vision tasks.
Current CIL algorithms can be divided into three groups.
The first group estimates the importance of each parameter and prevents important ones from being changed~\cite{aljundi2018memory,Afn}.
The second group utilizes knowledge distillation to maintain the model's discriminability~\cite{rebuffi2017icarl}.
Other methods rehearse former instances to overcome forgetting~\cite{zhao2020maintaining,zhu2021prototype,ISM,MgSvF,pretrain,SDC}.
\cite{pernici2021class} pre-allocates classifiers for future classes, which needs extra memory for feature tuning and is unsuitable for FSCIL.
Various approaches have been developed to address the challenge of retaining knowledge in incremental learning scenarios. iCaRL~\cite{rebuffi2017icarl}  employs replay and knowledge distillation to maintain previously learned knowledge. Other works explore different strategies such as saving embeddings instead of raw images, leveraging generative models for data rehearsal, task-wise adaptation, and output normalization to combat forgetting and adapt to new knowledge.

\noindent
\textbf{Few-Shot Class-Incremental Learning.} FSCIL addresses the dual challenges of FSL and CIL. Specifically, FSCIL focuses on learning from a minimal number of novel samples while retaining previously acquired knowledge. TOPIC~\cite{Topology} introduced the concept of FSCIL and utilized neural gas for topology preservation in the embedding space. Subsequent works ~\cite{MetaFSCIL} adapted existing CIL approaches to tackle FSCIL challenges. Other methods like ~\cite{cheraghian2021semantic} leverage word vectors to mitigate the intrinsic difficulty of data scarcity in FSCIL. An emerging approach involves meta-training on base class data, as seen in ~\cite{MetaFSCIL}, by simulating a number of fake incremental episodes for test scenarios. However, this often requires extra meta-training phases and parameter freezing, limiting practicality in real-world scenarios and the adaptability of models to novel concepts.
Indeed, while there has been significant progress in addressing forgetting and overfitting issues, achieving a unified framework to tackle both problems remains a challenge. The distribution calibration method~\cite{CalibratingCNN} introduced a promising approach to mitigate overfitting, but it faces limitations in scalability when applied to the context of FSCIL. Finding solutions that effectively combine both forgetting and overfitting mitigation in a scalable framework remains an active area of research.

\begin{figure*}[ht]
	\centering
	\includegraphics[width=1.\linewidth]{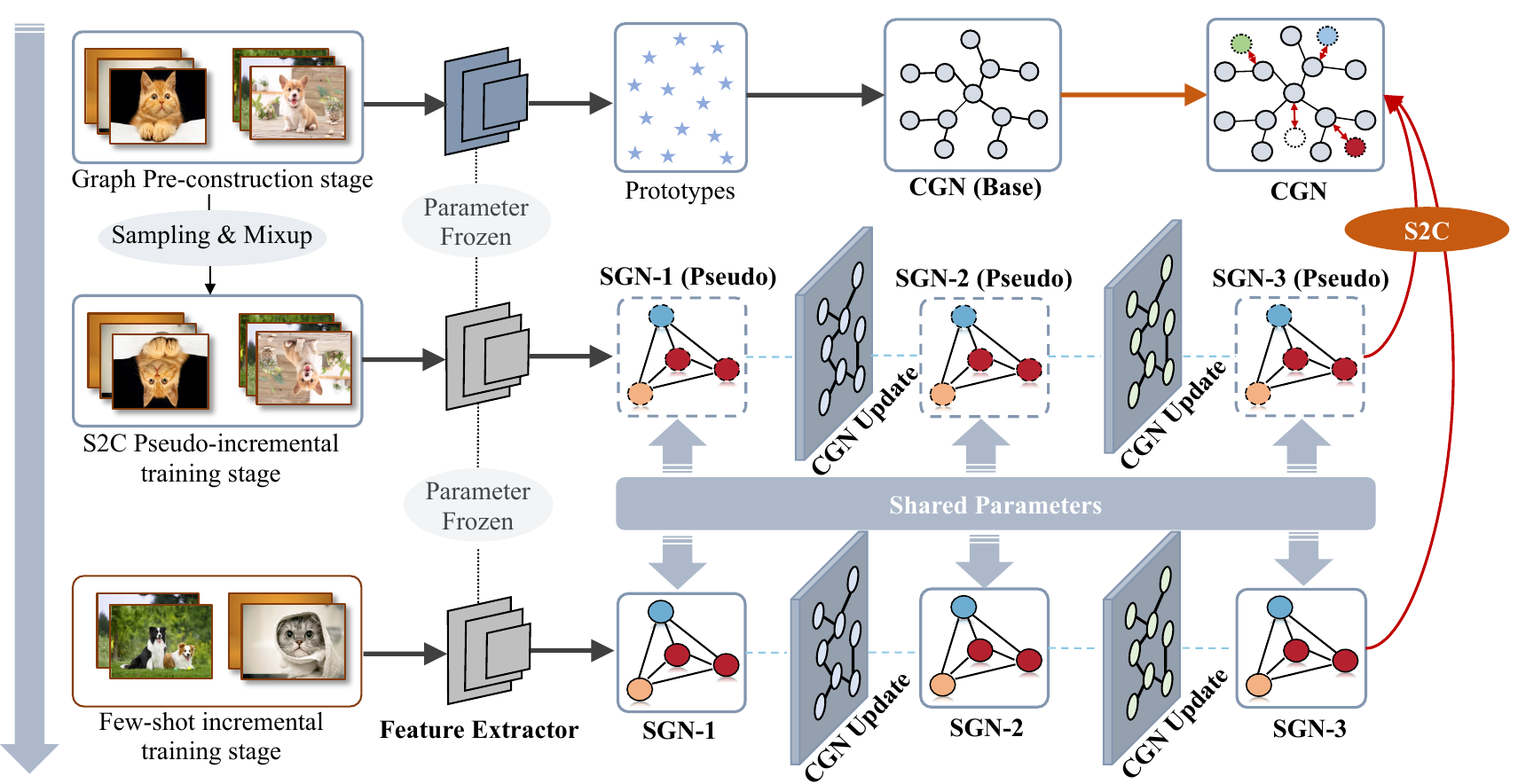}
	\caption{Our Sample-to-Class learning scheme for few-shot class-incremental learning. In the base session, we pre-train our feature extractor and construct the base class graph. In the Pseudo-incremental learning stage, we sythesize virtual tasks to make model fast adapt to few-shot scenario. }
	\label{fig:framework}
\end{figure*}

\section{Problem Description: FSCIL}
FSCIL has multiple continual tasks or sessions that appears in streams. 
Once the model starts to learn the current task, none of the previous data is available anymore. 
Besides, the evaluation of the model at each session involves the class in all previous sessions and current sessions. 
In concrete terms, given $T$ classification tasks with $\mathcal{D}_{\rm train}$  = \{$\mathcal{D}_{\rm train}^{t}$\}$_{t=0}^T$, where $\mathcal{D}_{\rm train}^{t}$ =\{($x_i$, $y_i$ )\}$_{i=0}^{NK}$ represents the training samples at session $t$. 
$x_i \in \mathcal{X}^t$ and $y_i \in \mathcal{Y}^t$ are the $i$-$th$ data and the corresponding label. 
We also denote $\mathcal{X}^t$ and $\mathcal{Y}^t$ as the sample set and label space at $t$-th session.
FSCIL task is to train a model from a continuous data stream in a class-incremental form, \ie, training sets $ \{  \mathcal{D}_{\rm train}^{0}, \mathcal{D}_{\rm train}^{1}, \dots \mathcal{D}_{\rm train}^{T} \} $.
The label set from different sessions are disjoint, \ie, $ \mathcal{Y}^{i} \bigcap \mathcal{Y}^{j} = \varnothing $ for $i \neq j$.
At the $t$-th learning session, only $\mathcal{D}_{\rm train}^{t}$ can be obtained for network training. When we step into the evaluation stage, the test dataset $ \mathcal{D}_{\rm test}^{t} $ should include test data from all classes that appears in previous and current sessions, \ie, all encountered label sets $ \{ \mathcal{Y}^{0}\cup \mathcal{Y}^{1}\dots \cup \mathcal{Y}^{t} \} $ at the $t$-th session.
For the first session, $ \mathcal{D}_{\rm train}^{0} $ has sufficient samples which is also called base training session.
For each class in the subsequent sessions, we have only a few samples . This training data is usually organized as a $N$-way $K$-shot, where $N$ denotes $N$ classes and $K$ denotes $K$ samples per class in dataset.
To measure an FSCIL model, we calculate the accuracy on the test set $ \mathcal{D}_{\rm test}^{t} $ at each session $ t $.

\section{Method}
In FSCIL, the number of each session is small, and the incremental training makes the old tasks forget.
Traditional FSL methods~\cite{EGNN} use GNN to establish relationships among few-shot samples, which effectively mitigate overfitting problems. 
Inspired by the use of GNN in FSL, we introduce GNN into FSCIL to create a sample-level graph that builds the underlying relationships among few-shot samples for each session. 
However, only the graph inside of each session is infeasible for the incremental scenario, because the previous samples are not available in the current few-shot training.
We seek to further establish dependencies among multiple classes from different sessions during the incremental learning process.
To this end, we introduce to build cross-session class-level graph on the basis of sample-level graph. As shown in Fig.~\ref{fig:framework}, 
given the two kinds of graphs, we also develop a novel Sample-to-Class (S2C) graph training strategy to leverage the deep relations in prediction. 
The framework includes sample-level and class-level graph networks, and leverage a multi-stage training strategy to improve the graph networks.


\begin{figure}[ht]
	\centering
	\includegraphics[width=1.\linewidth]{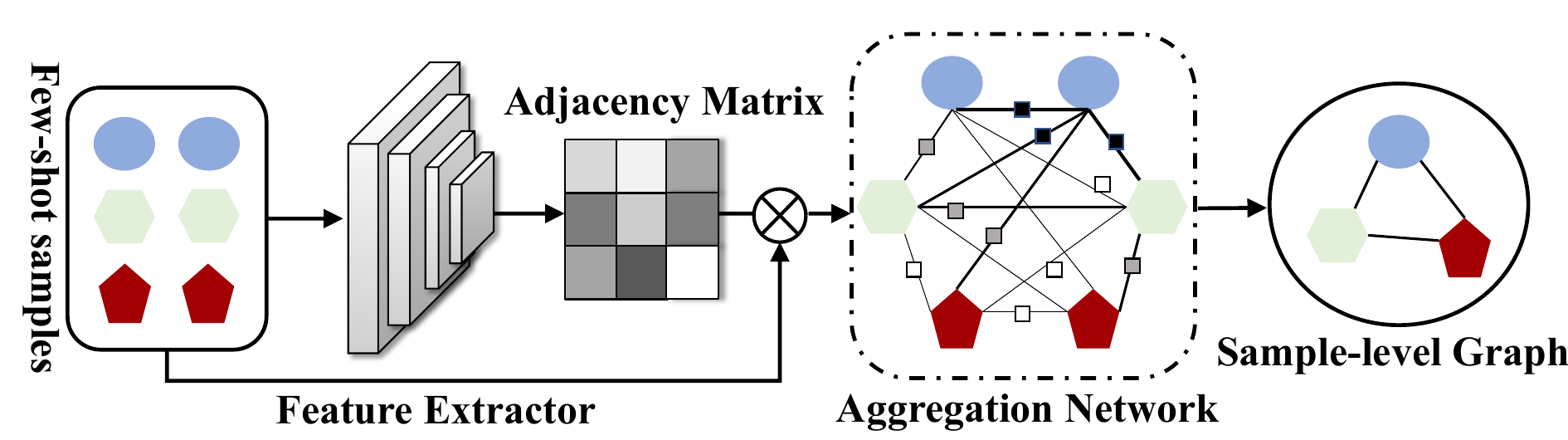}
	\caption{Sample-level Graph Neural Network. 
 }
	\label{fig:SGN}
\end{figure}

\subsection{Sample to Class (S2C) Graph Network}
\noindent
\textbf{4.1.1 Sample-Level Graph Network}

In traditional FSL, GNN is used to establish relationships between support and query samples. Inspired by this, we introduce the Sample-level Graph Network (SGN) to facilitate the learning of each FSL task. 
As shown in Fig.~\ref{fig:SGN}, for a current few-shot task, we first define the nodes of the SGN using the all available sample features belonging to different classes. 
Let $\mathcal{G}_{\rm SGN}$ = $\{ \mathcal{V}_{\rm SGN}, \mathcal{E}_{\rm SGN}\}$, where node set $\mathcal{V}_{\rm SGN}$ = $\{ \textbf{z}_1, \textbf{z}_2, \dots, \textbf{z}_k \}$ consists of the features $\textbf{z}$ of each sample.
The edge set $\mathcal{E}_{\rm SGN}$ of SGN is defined as relationship between nodes within each FSL task:
\begin{equation}
	e_{ij}^{\rm SGN} = \phi( \textbf{z}_{i} - \textbf{z}_{j}  ),
\end{equation} 
where $ \phi $ containing two Conv-BN-ReLU blocks, is the encoding network that transforms the instance similarity to a certain scale.

In this way, we construct a fully-connected sample-level graph based on the feature representations of all samples in the few-shot task. In the sample-level graph, each node corresponds to a feature, and each edge represents the relationship between the two connected nodes. By applying iterative aggregation operations of the GNN on both node information and edge information, the features of the samples are continuously updated, and the relationships between samples are re-established during this process. This allows for refined sample-level feature and a more accurate understanding of the relationships between samples.
Then, the obtained embeddings by SGN are averaged for each class as a refined class-level feature:
\begin{equation}
    \mathbf{p}_{c}^{\rm SGN} = \frac{1}{K}\sum\limits_{i=1}^{K}( \textbf{z}_{i} + \sum\limits_{j} (e_{ij}^{\rm SGN} \cdot \textbf{z}_{j})),
\end{equation}
where $\mathbf{p}_{c}^{SGN}$ represents the $c$-th refined class-level feature of few-shot task, $K$ is the number of samples in each class. 

In addition, to enhance the SGN model's capability to discover relationships between few-shot samples, we introduce the triplet loss into SGN:
\begin{equation}
        {L}_{\rm SGN} = \max(0, \lVert \textbf{z}_{i} - \textbf{z}_{P} \rVert^2 - \lVert \textbf{z}_{i} - \textbf{z}_{N} \rVert^2 + m),
\end{equation}
where $m$ is a margin parameter which can be used to control the distance between positive and negative samples. $z_{P}$ and $z_{N}$ is represented as the features of positive samples and negative samples respectively.
This loss function is designed to increase the distance between samples from the same class while simultaneously decreasing the distance between samples from different classes. This strategy aims to improve the discriminative power of the SGN in distinguishing between samples and effectively capturing sample-level relationships.

After SGN in-depth exploration of the relationships among the few-shot samples, we obtain the class-level features of the most representative few-shot classes. 
However, SGN can only assess sample-level relationships within a few-shot session.
That is, when a new session begins, the relationships of the old samples cannot be used in the current training, yielding catastrophic forgetting.
Motivate by this, we try to establish class-level relationships among multiple few-shot sessions. 

~

\noindent
\textbf{4.1.2 Class-Level Graph Network}

The relationship established by SGN is limited to the samples within a same session and cannot be established for class-level features under different sessions. 
In other words, the model need to adapt to new FSL tasks while simultaneously retaining proficiency in previously encountered tasks. 
To this end, we use class-level features as a medium to form dependencies between old and new classes, and construct Class-level Graph Network (CGN) in the incremental learning scenarios.
CGN leverages previously learned knowledge to aid in the learning of the current few-shot task, allowing for more robust and efficient learning across multiple sessions.

\begin{figure}[ht]
	\centering
	\includegraphics[width=1.\linewidth]{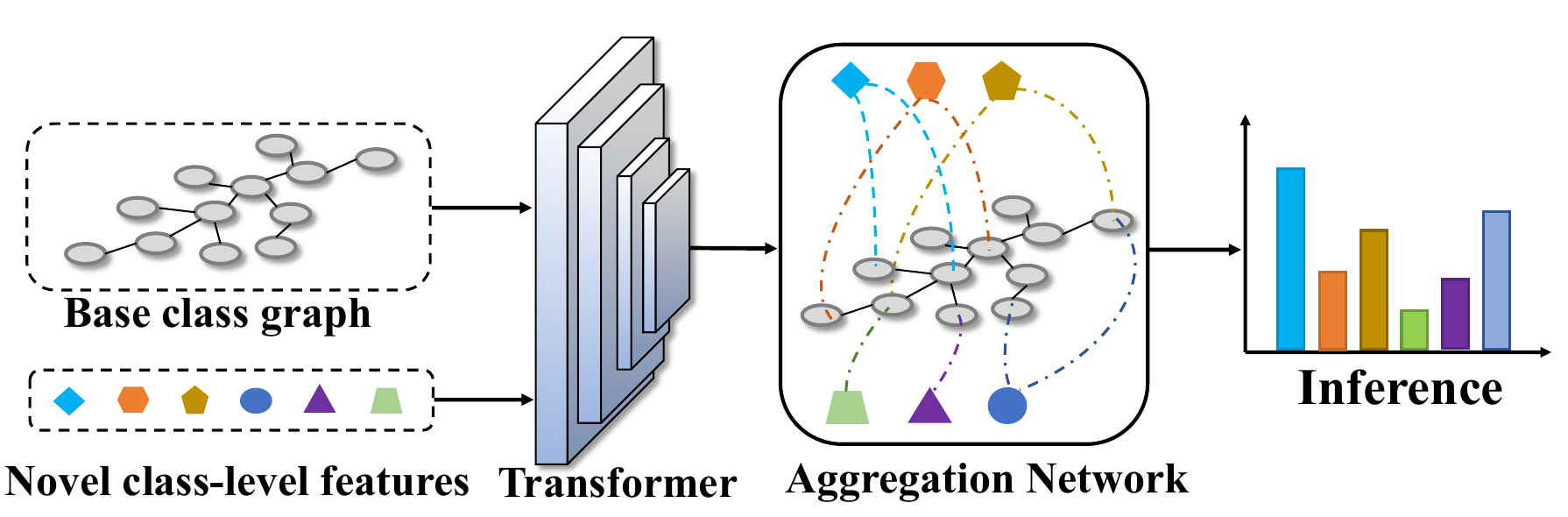}
	\caption{Class-level Graph Neural Network.
 }
	\label{fig:CGN}
\end{figure}

As shown in Fig.~\ref{fig:CGN}, in CGN, we combine the Transformer~\cite{vaswani2017attention} with the GNN to build links between novel and old classes by utilizing the precise capture of global information. 
Specifically, the base graph and the refined class-level features exported by SGN are used as input to the CGN. Then, we use the multi-head attention mechanism to construct the relationship between the old class and new class, and use the GNN to aggregate these information to iteratively calibrate the prototypes of the novel class. Eventually a class-level feature graph with well-established relationships is outputted.
We set the parameters query $ \mathbf{q} $, key $ \mathbf{k} $ and value $ \mathbf{v} $ to
\begin{equation}
\begin{aligned}
    \mathbf{v} = \mathbf{p}_{c}^{\rm SGN}, \quad
    \mathbf{k} = W_{k}^{T} \mathbf{v}, \quad
    \mathbf{q} = W_{q}^{T} \mathbf{v},
\end{aligned}
\end{equation}
$ W_{k}$ and $ W_{q} $ are the learnable parameter of linear projection function. 
The class-level features formula after CGN calibrating operation is as follows:
\begin{equation}
	\mathbf{p}^{\rm CGN}_{c} = \mathbf{p}_{c}^{\rm SGN} + \frac{\mathbf{k}^{T} \mathbf{q}}{\sqrt{d} }\mathbf{v},
\end{equation}
where $\sqrt{d}$ is a scaled factor. 
To keep the distinction between the new class and the old class, we define the following per-sample loss function to learn CGN:
\begin{equation}
	L_{\rm CGN} = L\left(G\left[cos(\mathbf{z}_i, \mathbf{p}^{\rm CGN}_{c})\right],y_i\right). 
\end{equation}

\noindent
\textbf{4.1.3 S2C loss function}

S2C is trained by optimizing the following loss function:
\begin{equation}
    L = L_{\rm SGN} + \alpha L_{\rm CGN},
\end{equation}
where $\alpha$ is a pre-defined scaled factor.

With the help of the SGN, the CGN connects class-level features with the rich semantic information obtained from SGN. 
CGN establishes connections between class-level features from all sessions through an attention mechanism, resulting in a graph with abundant class-level features. 
This graph is then used for subsequent label prediction tasks, enhancing the model's ability to make predictions.


\subsection{S2C Training Procedure for FSCIL}

Nevertheless, it is still difficult to build S2C graph, because of the very small number of samples for each session.
In FSCIL, before the few-shot incremental sessions, a base session is used for pre-training the model~\cite{zhang2021few}.
In the base session, there are an ample number of training instances available to build the initial model. 
Inspired by the meta learning~\cite{Metalearning}, we propose to pre-learn how to build graph from sample-level to class-level within the base session.
Specifically, as shown in Fig.~\ref{fig:framework}, we design a multi-stage training strategy for S2C. The strategy consists of three stages, namely Graph pre-construction stage, S2C pseudo-incremental training stage and Few-shot incremental training stage.


\noindent
\textbf{4.2.1 Graph pre-construction stage}

Before few-shot sessions, the base session offers a substantial volume of data that can serve as prior knowledge for the model to tackle subsequent few-shot tasks, thereby helping to alleviate the overfitting issue. 
Nevertheless, this prior knowledge is often underutilized and doesn't effectively aid in learning subsequent knowledge, creating a significant hindrance to FSCIL. 
To tackle this problem, we employ a strategy to compute class-level features enriched with semantic knowledge by extracting features from a substantial number of samples. A base graph is built based on the similarity relationships between these class-level features, which can be updated and adapted to subsequernt tasks.


Specifically, we first pretrain a feature extractor in the base session, using training samples from $\mathcal{D}_{\rm train}^{0}$:
\begin{equation}
	\theta^{\ast} = \min\limits_{\theta} \mathcal{L}\left( G\left[f_{\theta}(x)\right],y\right),
\end{equation}
where $ \mathcal{L}(\cdot) $ represents cross-entropy loss function, $f_{\theta}(\cdot)$ is the feature extractor parameterized by $\theta$ and $G(\cdot)$ denotes the classifier. 
Let $\mathcal{G}_{\rm base}=\{\mathcal{V}, \mathcal{E}\}$ denote the base graph, where $\mathcal{V}_{\rm base}$ = $\{ \mathbf{v}_1, \mathbf{v}_2, \dots, \mathbf{v}_M \}$
is the node set and $\mathcal{E}_{\rm base}$ is the edge set. 
In the base graph, we first initiate the nodes with base class prototype:
\begin{equation}
	\mathbf{v}_m = \frac{1}{N}\sum\limits_{n=1}^{\left |D_\mathrm{train}^{0} \right|} f_\theta(x_n) \cdot \mathbb{I}(y_m=y_n),
\end{equation}
where $N$ is the number of samples belonging to the $m$-$th$ class and $\mathbb{I}(\cdot)$ is the indicator function. Then, our base graph edges $\mathcal{E}$ is defined as similarity between nodes $\mathbf{v}_m$ and $\mathbf{v}_n$ :
\begin{equation}
    e_{mn} = \frac{\mathbf{v}_m^{\rm T} \mathbf{v}_n}{\Vert \mathbf{v}_m \Vert  \Vert \mathbf{v}_n \Vert}.
\end{equation}
Establishing base graph lays the foundation for subsequent incremental class learning. The base graph not only provide prior knowledge for the learning of new classes but also serve as a medium for connecting SGN to CGN.

\noindent
\textbf{4.2.2 S2C pseudo-incremental training stage}

In order to enhance S2C model's capability to learn from few-shot data, we design to make model learn how to construct graphs in FSCIL scenarios ahead of time.
To this end, we devise the pseudo-incremental learning process. This process operates within the base session and is tailored to bolster the model's capacity to effectively adapt to new FSL tasks. 
To enhance the model's discriminative ability for new classes in forthcoming tasks, we introduce a \textit{meta-learning-based pseudo-incremental training paradigm}. This paradigm equips the model with the skills to learn how to effectively grasp a new class using only a few samples.
Specifically, we stochastically draw $N$ FSL tasks, denoted as $T_{1}$ to $T_{N}$ from the training set $\mathcal{D}_{\rm train}^{0}$.
These tasks are characterized by an $N$-way $K$-shot setup, satisfying the condition $\mathcal{Y}^{1} \cap \mathcal{Y}^{2} \cap \dots \mathcal{Y}^{n}  = \varnothing$. 
Note that these FSL tasks serve as foundational tasks within the pseudo-incremental process.

Moreover, we employ manifold mixup ~\cite{verma2019manifold} to fuse instances, treating the resulting fused instances as virtual incremental classes. 
We fuse two samples from different FSL tasks to generate new virtual samples $\textbf{z}$ which serve as data for virtual task $ \mathcal{T} $:
\begin{equation}
	\textbf{z} = \sum\limits_{i}^{NK} \lambda f_{\theta}(x_{i}^{t_{1}}) + (1 - \lambda)f_{\theta}(x_{i}^{t_{2}}),
\end{equation}
where $ \lambda \in [0,1] $ is sampled from Beta distribution, 
and $ \textbf{z} $ represents the feature of the sample in FSL task.
Superscript $t_1$ and $t_2$ denotes different tasks. 
In this way, we strive to imbue the model with enhanced proficiency in assimilating and adapting to new knowledge in the FSCIL context. 
The pseudo-incremental learning paradigm enables S2C model to achieve the capility of building graph relationships among samples and classes before few-shot sessions. 
In the following subsections, we introduce how to build sample-level to class-level graph in the FSCIL process.

\noindent
\textbf{4.2.3 Few-shot incremental training stage} 

Once the feature backbone is stabilized during the base session, and both the SGN and CGN have been trained in the S2C adaptation stage, our S2C model is ready to be applied to the task of few-shot class-incremental learning. 
In the subsequent stages, we feed the novel few-shot data into the pre-trained SGN, which update the nodes within the CGN. During the prediction phase, we utilize a metric-based evaluation approach to make predictions regarding the labels of the query nodes.

In S2C, SGN (see Fig.~\ref{fig:SGN}) is built to analyze the relationship of a few samples to aggregate similar samples and obtains refined class-level features. SGN matches the class-level features after learning with the base graph,  
which not only strengthens SGN's ability to learn FSL tasks but also reduces the interference to other classes. 
CGN (see Fig.~\ref{fig:CGN}) extends the calibrated class-level features to the base class graph and predicts the label of query samples. 
With the full cooperation of SGN and CGN, our S2C model learns more representative features while construct the links between multiple classes from different sessions.

\noindent
\textbf{4.2.4 Discussion} 

In the multi-stage training process of S2C, we initially build the base graph to preserve the knowledge from the base dataset, which could aid in subsequent class-incremental learning. 
Then, we conducted a S2C adaptation stage, allowing the S2C model to adapt to the few-shot data beforehand. Finally, we deployed the S2C model in the real FSCIL tasks. This multi-stage approach enables the S2C model to perform effectively in FSCIL.

 
In general, we introduce the S2C model for FSCIL which comprise two essential components: SGN and CGN. S2C is designed to establish feature dependencies among various sessions based on both sample-level and class-level features. We have also outlined a multi-stage training strategy for S2C, which enables the model to be effectively deployed in FSCIL tasks.

\begin{table*}[ht]
        \scriptsize
	\centering
	\caption{Comparison with the state-of-the-art on CIFAR100 dataset.}\label{lab:cifar}
	\resizebox{1.\linewidth}{!}{
		\begin{tabular}{l cccccccccc cc  }
			\bottomrule
			\multirow{2}{*}{\textbf{Methods}} &\multicolumn{9}{c}{\textbf{Accuracy in each session(\%)}}  & \multirow{2}{*}{\textbf{PD}} & \multirow{2}{*}{\textbf{Improve}}\\
			\cline{2-10}
			&\textbf{0} &\textbf{1}  &\textbf{2}&\textbf{3} &\textbf{4}&\textbf{5}  &\textbf{6}&\textbf{7}&\textbf{8} & &\\ 
			\hline														
			Finetune           &  64.10  &  39.61  &  15.37  &   9.80  &   6.67  &   3.80  &   3.70  &   3.14  &   2.65  &  61.45  &  \textbf{+38.66}  \\
			iCaRL~\cite{rebuffi2017icarl}            &  64.10  &  53.38  &  41.69  &  34.13  &  27.93  &  25.06  &  20.41  &  15.48  &  13.73  &  50.37  &  \textbf{+22.85}  \\
			EEIL~\cite{castro2018end}              &  64.10  &  53.11  &  43.71  &  35.15  &  28.96  &  24.98  &  21.01  &  17.26  &  15.85  &  48.25  & \textbf{+20.48}   \\
			Rebalancing~\cite{hou2019learning}        &  64.10  &  53.05  &  43.96  &  36.97  &  31.61  &  26.73  &  21.23  &  16.78  &  13.54  &  50.56  &  \textbf{+25.89}  \\
			TOPIC~\cite{tao2020few}              &  64.10  &  55.88  &  47.07  &  45.16  &  40.11  &  36.38  &  33.96  &  31.55  &  29.37  &  34.73  &  \textbf{+15.64}  \\
			Decoupled-Cosine~\cite{vinyals2016matching}   &  74.55  &  67.43  &  63.63  &  59.55  &  56.11  &  53.80  &  51.68  &  49.67  &  47.68  &  26.87  &  \textbf{+3.49}  \\
			Decoupled-DeepEMD~\cite{zhang2020deepemd}  &  69.75  &  65.06  &  61.20  &  57.21  &  53.88  &  51.40  &  48.80  &  46.84  &  44.41  &  25.34  &  \textbf{+5.11}  \\
			F2M~\cite{shi2021overcoming}                &  64.71  &  62.05  &  59.01  &  55.58  &  52.55  &  49.96  &  48.08  &  46.28  &  44.67  &  20.04  &  \textbf{+1.38}  \\
			CEC~\cite{zhang2021few}                &  73.07  &  68.88  &  65.26  &  61.19  &  58.09  &  55.57  &  53.22  &  51.34  &  49.14  &  23.93  &  \textbf{+3.12}  \\
			LIMIT~\cite{zhou2022few}             &  73.81  &  72.09  &  67.87  &  63.89  &  60.70  &  57.77  &  55.67  &  53.52  &  51.23  &  22.58  &  \textbf{+1.85}  \\
			FACT~\cite{zhou2022forward}               &  74.60  &  72.09  &  67.56  &  63.52  &  61.38  &  58.36  &  56.28  &  54.24  &  52.10  &  22.50  &  \textbf{+0.82}  \\
                MCNet~\cite{Ji2023MCNet}  &73.30  &69.34 &65.72  &61.70  &58.75  &56.44  &54.59  &53.01  &50.72  &22.58  & \textbf{+0.62}     \\
                MFS3~\cite{Xu2023MFS3}  &73.42  &69.85  &66.44  &62.81  &59.78  &56.94  &55.04  &53.00  &51.07  &22.35  & \textbf{+0.39} \\
			\hline
			S2C(ours)       	   &  \textbf{75.15}  &  \textbf{73.07}  &  \textbf{68.31}  &  \textbf{64.61}  &  \textbf{61.94}  &  \textbf{59.41}  &  \textbf{57.62}  &  \textbf{55.62}  &  \textbf{53.19}  &  \textbf{21.96}  &     \\
			\toprule																
	\end{tabular}}
\end{table*}

\begin{table*}[ht]
        \scriptsize
	\centering
	\caption{Comparison with the state-of-the-art on MiniImageNet dataset.}\label{lab:mini}
	\resizebox{1.\linewidth}{!}{
		\begin{tabular}{l cccccccccc cc  }
			\bottomrule
			\multirow{2}{*}{\textbf{Methods}} &\multicolumn{9}{c}{\textbf{Accuracy in each session(\%)}}  & \multirow{2}{*}{\textbf{PD}} & \multirow{2}{*}{\textbf{Improve}}\\
			\cline{2-10}
			&\textbf{0} &\textbf{1}  &\textbf{2}&\textbf{3} &\textbf{4}&\textbf{5}  &\textbf{6}&\textbf{7}&\textbf{8} & &\\ 
			\hline														
			Finetune           &  61.31  &  27.22  &  16.37  &   6.08  &   2.54  &   1.56  &   1.93  &   2.60  &   1.40  &  59.91  &  \textbf{+38.66}  \\
			iCaRL~\cite{rebuffi2017icarl}            &  61.31  &  46.32  &  42.94  &  37.63  &  30.49  &  24.00  &  20.89  &  18.80  &  17.21  &  44.10  &  \textbf{+22.85}  \\
			EEIL~\cite{castro2018end}              &  61.31  &  46.58  &  44.00  &  37.29  &  33.14  &  27.12  &  24.10  &  21.57  &  19.58  &  41.73  & \textbf{+20.48}   \\
			Rebalancing~\cite{hou2019learning}        &  61.31  &  47.80  &  39.31  &  31.91  &  25.68  &  21.35  &  18.67  &  17.24  &  14.17  &  47.14  &  \textbf{+25.89}  \\
			TOPIC~\cite{tao2020few}              &  61.31  &  50.09  &  45.17  &  41.16  &  37.48  &  35.52  &  32.19  &  29.46  &  24.42  &  36.89  &  \textbf{+15.64}  \\
			Decoupled-Cosine~\cite{vinyals2016matching}   &  70.37  &  65.45  &  61.41  &  58.00  &  54.81  &  51.89  &  49.10  &  47.27  &  45.63  &  24.74  &  \textbf{+3.49}  \\
			Decoupled-DeepEMD~\cite{zhang2020deepemd}  &  69.77  &  64.59  &  60.21  &  56.63  &  53.16  &  50.13  &  47.79  &  45.42  &  43.41  &  26.36  &  \textbf{+5.11}  \\
			F2M~\cite{shi2021overcoming}                &  67.28  &  63.80  &  60.38  &  57.06  &  54.08  &  51.39  &  48.82  &  46.58  &  44.65  &  22.63  &  \textbf{+1.38}  \\
			CEC~\cite{zhang2021few}                &  72.00  &  66.83  &  62.97  &  59.43  &  56.70  &  53.73  &  51.19  &  49.24  &  47.63  &  24.37  &  \textbf{+3.12}  \\
			LIMIT~\cite{zhou2022few}             &  72.32  &  68.47  &  64.30  &  60.78  &  57.95  &  55.07  &  52.70  &  50.72  &  49.19  &  23.13  &  \textbf{+1.85}  \\
			FACT~\cite{zhou2022forward}               &  72.56  &  69.63  &  66.38  &  62.77  &  60.60  &  57.33  &  54.34  &  52.16  &  50.49  &  22.07  &  \textbf{+0.82}  \\
                MCNet~\cite{Ji2023MCNet}  &72.33  &67.70 &63.50  &60.34  &57.59  &54.70  &52.13  &50.41  &49.08  &23.25  & \textbf{+2.00}     \\
                MFS3~\cite{Xu2023MFS3}  &\textbf{73.65}  &68.91  &64.60  &61.48  &58.68  &55.55  &53.33  &51.69  &50.26  &23.39  & \textbf{+2.14} \\
			\hline
			S2C(ours)       	   &  73.25  &  \textbf{71.57}  &  \textbf{67.46}  &  \textbf{64.01}  &  \textbf{61.04}  &  \textbf{58.41}  &  \textbf{55.62}  &  \textbf{53.62}  &  \textbf{52.00}  &  \textbf{21.25}     \\
			\toprule																
	\end{tabular}}
\end{table*}

\section{Experiment}
\subsection{Dateset}

We evaluate the effectiveness of the proposed method on datasets MiniImageNet, CUB200-2011 and CIFAR100. 
\begin{itemize}
    \item \textit{MiniImageNet}~\cite{MiniImagenet} is a subset of the ImageNet dataset,  specifically designed for evaluating models’ performance in scenarios where only a limited number of examples are available for each class. MiniImageNet contains 100 classes, each with 600 color images of size 84$\times$84 pixels.
    \item \textit{CIFAR100}~\cite{cifar} consists of 100 classes, each representing a different object category. The dataset contains 6,000 32$\times$32 RGB images, with 600 images per class.
    \item \textit{Caltech-UCSD Birds-200-2011}~\cite{Cub} CUB-200 is a widely used benchmark dataset in the field of fine-grained bird species recognition. 
    The dataset contains 200 different bird species, each of which is with a set of annotated images. 
    The dataset consists of 11,788 images in total. 
\end{itemize}
For MiniImageNet and CIFAR100, 100 classes are divided into 60 base classes and 40 new classes.  
The new classes are formulated into eight 5-way 5-shot incremental tasks. 
For CUB200, 200 classes are divided into 100 base classes and 100 incremental classes, and the new classes are formulated into ten 10-way 5-shot incremental tasks. 

\begin{table*}[ht]
\scriptsize
\centering
\caption{Comparison with the state-of-the-art on CUB200 dataset.}\label{lab:CUB}
\resizebox{1\linewidth}{!}{
 \begin{tabular}{l cccccccccccc cc  }
  \toprule
  \multirow{2}{*}{\textbf{Method}} &\multicolumn{11}{c}{\textbf{Accuracy in each session(\%)}}  & \multirow{2}{*}{\textbf{PD}} & \multirow{2}{*}{\textbf{Improve}}\\
  \cline{2-12}
  & \textbf{0} &\textbf{1}  &\textbf{2}&\textbf{3} &\textbf{4}&\textbf{5}  &\textbf{6}&\textbf{7}&\textbf{8}&\textbf{9}&\textbf{10}  & & \\ 
  \hline              
  Finetune           &  68.68  &  43.70  &  25.05  &  17.72  &  18.08  &  16.95  &  15.10  &  10.06  &   8.93  &   8.93  &   8.47  &  60.21  &  \textbf{+41.75}  \\
  iCaRL~\cite{rebuffi2017icarl}              &  68.68  &  52.65  &  48.61  &  44.16  &  36.62  &  29.52  &  27.83  &  26.26  &  24.01  &  23.89  &  21.16  &  47.52  &  \textbf{+29.06}  \\
  EEIL~\cite{castro2018end}               &  68.68  &  53.63  &  47.91  &  44.20  &  36.30  &  27.46  &  25.93  &  24.70  &  23.95  &  24.13  &  22.11  &  46.57  &  \textbf{+28.11}  \\
  Rebalancing~\cite{hou2019learning}         &  68.68  &  57.12  &  44.21  &  28.78  &  26.71  &  25.66  &  24.62  &  21.52  &  20.12  &  20.06  &  19.87  &  48.81  &  \textbf{+30.35}  \\
  TOPIC~\cite{tao2020few}              &  68.68  &  62.49  &  54.81  &  49.99  &  45.25  &  41.40  &  38.35  &  35.36  &  32.22  &  28.31  &  26.26  &  42.40  &  \textbf{+23.94}  \\
  SPPR~\cite{zhu2021self}               &  68.68  &  61.85  &  57.43  &  52.68  &  50.19  &  46.88  &  44.65  &  43.07  &  40.17  &  39.63  &  37.33  &  31.35  &  \textbf{+12.89}  \\
  Decoupled-NetCosine&  74.96  &  70.57  &  66.62  &  61.32  &  60.09  &  56.06  &  55.03  &  52.78  &  51.50  &  50.08  &  48.47  &  26.49  &  \textbf{+8.03}  \\
  Decoupled-Cosine~\cite{vinyals2016matching}   &  75.52  &  70.95  &  66.46  &  61.20  &  60.86  &  56.88  &  55.40  &  53.49  &  51.94  &  50.93  &  49.31  &  26.21  &  \textbf{+7.75}  \\
  Decoupled-DeepEMD~\cite{zhang2020deepemd}  &  75.35  &  70.69  &  66.68  &  62.34  &  59.76  &  56.54  &  54.61  &  52.52  &  50.73  &  49.20  &  47.60  &  27.75  &  \textbf{+9.29}  \\
  CEC~\cite{zhang2021few}                &  75.85  &  71.94  &  68.50  &  63.50  &  62.43  &  58.27  &  57.73  &  55.81  &  54.83   & 53.52 & 52.28  &  23.57  &  \textbf{+5.11} \\
  FACT~\cite{zhou2022forward}                &  75.90  &  73.23  &  70.84  &  66.13  &  65.56  &  62.15  &  61.74  &  59.83  &  58.41   & 57.89 & 56.94  &  18.96  &  \textbf{+0.50} \\
  MCNet~\cite{Ji2023MCNet}                &  75.90  &  73.23  &  70.84  &  66.13  &  65.56  &  62.15  &  61.74  &  59.83  &  58.41   & 57.89 & 56.94  &  18.96  &  \textbf{+0.50} \\
  MFS3~\cite{Xu2023MFS3}                &  75.63  &  72.51  &  69.65  &  65.29  &  63.13  &  60.38  &  58.99  &  57.41  &  55.55   & 54.95 & 53.47  &  22.16  &  \textbf{+3.70} \\
  \hline
  S2C(ours)   &  \textbf{75.92}  &  \textbf{73.57}  &  \textbf{71.67}  &  \textbf{68.01}  &  \textbf{66.94}  &  \textbf{63.61}  &  \textbf{62.22}  &  \textbf{61.42}  &  \textbf{59.79}   & \textbf{58.56} & \textbf{57.46}  &  \textbf{18.46}  \\
  \toprule
	\end{tabular}}
\end{table*}

\noindent
\subsection{Training and evaluation protocol}
For CIFAR100, we use ResNet20, while for other datasets we use ResNet18. We optimize with stochastic gradient descent using momentum 0.9, and the learning rate is set to 0.1 and decays with cosine annealing. 
We evaluate models after each session on the test set $ \mathcal{D}_{\rm test}$ and report the Top 1 accuracy. 
We also use a performance dropping rate ({PD}) that measures the absolute accuracy drops in the last session w.r.t.
the accuracy in the first session, \ie, $ \textrm{PD} =  {A}_{0} - {A}_{N} $, 
where $ {A}_{0} $ is the classification accuracy of the base session and $ {A}_{N} $ is the accuracy of the last session.

\begin{figure*}[ht]
	\centering
	\includegraphics[width=1.\linewidth]{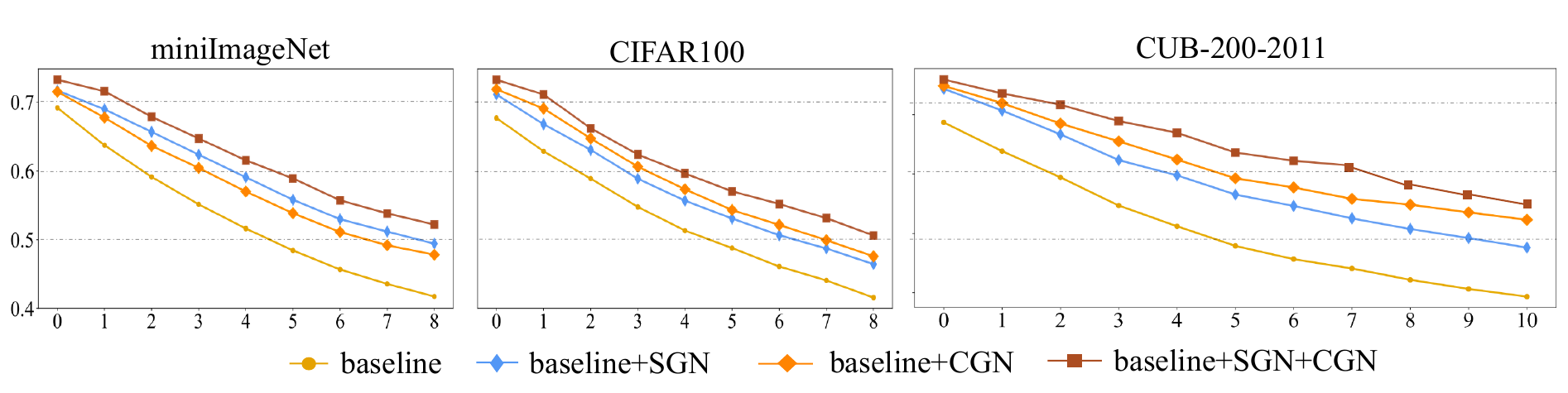}
	\caption{Ablation studies on MiniImageNet, CIFAR100 and CUB-200-2011. 
 }
	\label{fig:ablation}
\end{figure*}

\begin{figure}[ht]
	\centering
	\includegraphics[width=1.\linewidth]{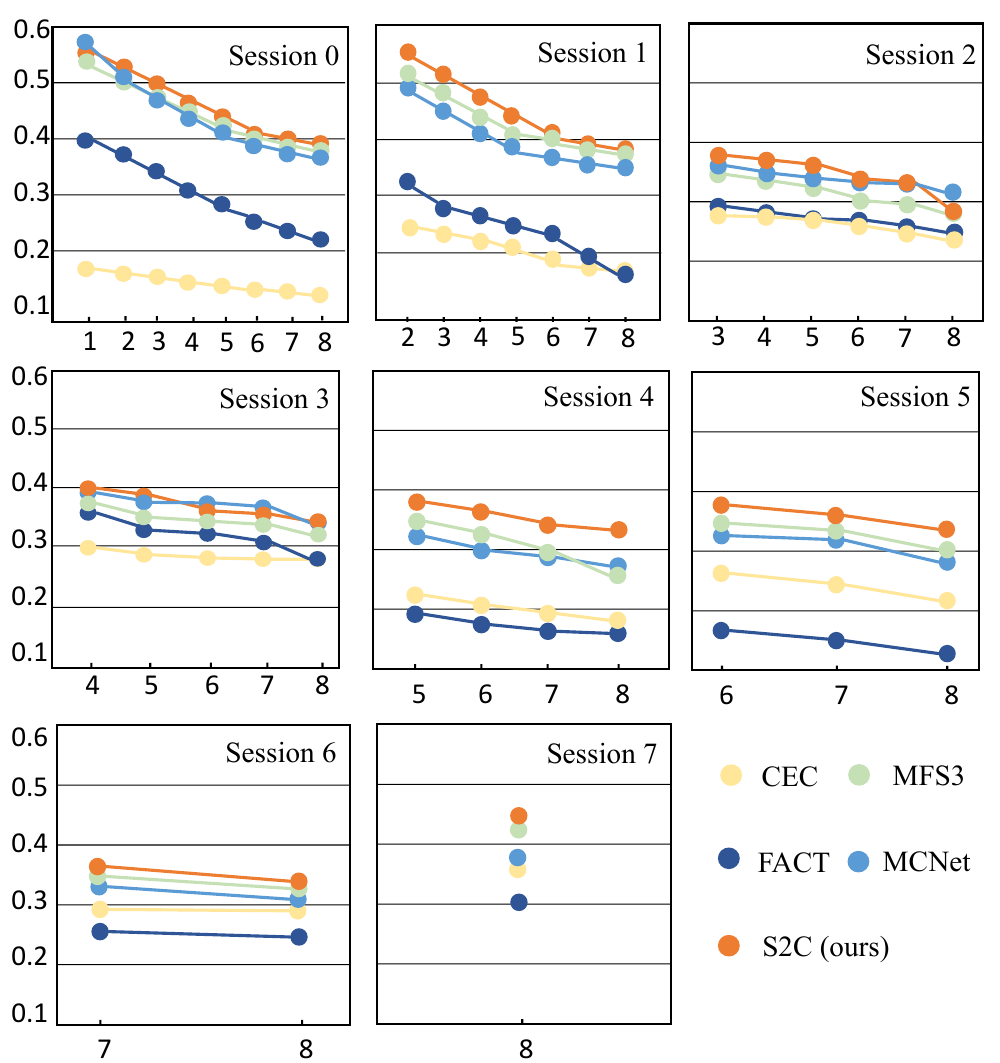}
	\caption{The accuracy of each session in MiniImageNet dataset in the FSCIL task learning process.
 }
	\label{fig:task_figures}
\end{figure}

\begin{figure}[ht]
	\centering
	\includegraphics[width=1\linewidth]{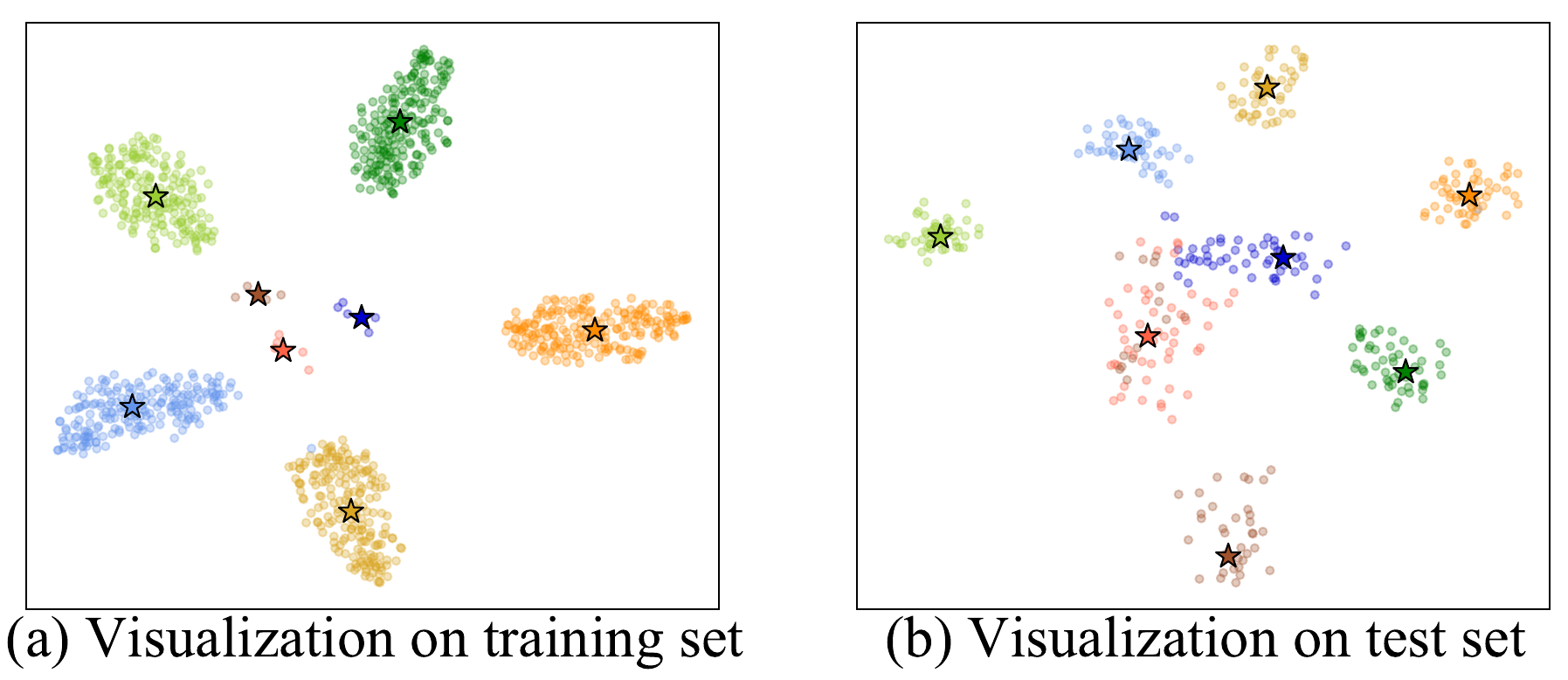}
	\caption{Visualization of decision boundary of training set and test set on CUB200-2011.Circles represent sample features, stars represent class-level features, and different colors represent different categories.}
	\label{fig:tsne}
\end{figure}

\subsection{Training details }
We adhere to standard data preprocessing and augmentation protocols, encompassing random resizing, random flipping, and color jittering. Our model training employs a batch size of 512 during the base session, and a batch size of 128 in each incremental session. On the miniImageNet dataset, the base session spans 500 epochs, with each incremental session spanning 100 iterations. Initial learning rates stand at 0.1 for the base session and 0.05 for incremental sessions. For CIFAR-100, we conduct 300 epochs in the base session, with each incremental session spanning 100 iterations. Initial learning rates remain consistent at 0.1 for both base and incremental sessions. On the CUB-200 dataset, we train for 100 epochs during the base session, and each incremental session covers 80 iterations. Initial learning rates remain consistent at 0.1 for the base session and 0.05 for incremental sessions. Across all experiments, a cosine annealing strategy governs the learning rate, and the optimizer utilized is SGD with momentum 0.9. 
The top-1 accuracy and performance dropping (forgetting) rate is introduced to evaluate models after each session.

\subsection{Major comparison}
We compare our proposed S2C method with existing methods and report the performance on three FSCIL benchmark datasets in Tables~\ref{lab:cifar}, ~\ref{lab:mini} and ~\ref{lab:CUB}. These methods include classical CIL methods, such as iCaRL~\cite{rebuffi2017icarl}, EEIL~\cite{castro2018end}, and Rebalancing~\cite{hou2019learning}, as well as continual-trainable FSCIL methods like TOPIC~\cite{tao2020few}, and backbone-frozen FSCIL methods such as SPPR~\cite{zhu2021self}, DeepEMD/Cosine/NegCosine~\cite{liu2020negative,vinyals2016matching,zhang2020deepemd},  CEC~\cite{zhang2021few}, and FACT~\cite{zhou2022forward} and model-complement methods such as MCNet~\cite{Ji2023MCNet}, MFS3~\cite{Xu2023MFS3}. 
We also include a simple baseline, labeled as 'finetune', where the model is directly fine-tuned using the limited available data.
As the whole, we observe that S2C consistently outperforms the current SOTA method on benchmark datasets. 
The performance of S2C method is higher than that of other methods, and the performance dropping rate is lower than that of other methods. 
Specifically, our PD outperforms the SOTA results by 0.39 on CIFAR100, 0.82 on miniImageNet and 0.50 on CUB200.
The poor performance of CIL method (such as iCaRL) indicates that classical CIL methods primarily focus on extending the model with sufficient instances and are not well-suited for few-shot task.
S2C has better performance than Decoupled-DeepEMD/Cosine/NegCosine~\cite{liu2020negative,vinyals2016matching,zhang2020deepemd}, CEC~\cite{zhang2021few} and FACT~\cite{zhou2022forward}, MCNet~\cite{Ji2023MCNet} and MFS3~\cite{Xu2023MFS3}. 
It reveals that in FSCIL, continual-trainable FSCIL methods encounter overfitting issues and perform poorly in incremental sessions, it is important to make FSL tasks be trained well which strengthens new task constraints to reduce the impact on old tasks. 

As shown in Fig.~\ref{fig:task_figures}, we compared the accuracy of each session on the MiniImageNet dataset with the CEC~\cite{zhang2021few}, FACT~\cite{zhou2022forward}, MCNet~\cite{Ji2023MCNet} and MFS3~\cite{Xu2023MFS3}methods. 
It can be seen from the figure that in the FSCIL task learning process, the performance of each session is higher than that of other methods. 

\subsection{Ablation Study}
We conducted an in-depth analysis of the significance of each component within the S2C approach on datasets MiniImageNet, CIFAR100, and CUB-200-2011. 
The results are presented in Fig.~\ref{fig:ablation}. We designed models with varying combinations of core S2C elements for comparison.
The ``Baseline'' model denotes the scenario where the backbone network directly learns FSCIL tasks. By examining Fig.~\ref{fig:ablation}, we deduce the following insights:
1) The incorporation of the CGN module effectively mitigates the issue of catastrophic forgetting that is observed in the baseline model during FSCIL tasks.
2) The integration of the SGN module elevates the learning performance of FSL tasks. This enhancement is reflected not only in FSL tasks but also overall across sessions, highlighting the significance of SCN for FSL task training.
3) Combining both SGN and CGN modules not only enhances FSL task performance but also takes into consideration semantic conflicts arising due to data imbalance and other factors between old and new classes.
Through ablation experiments, we establish that both the SGN and CGN modules significantly contribute to the success of FSCIL tasks. 

\subsection{Visualization of Incremental Session}
We visually represent the learned decision boundaries using t-SNE on the CUB-200-2011 dataset, as depicted in Fig~\ref{fig:tsne}: 
1) Fig.~\ref{fig:tsne}(a): This panel illustrates the decision boundary of the training set, where we trained on five old classes and three new classes with a limited number of samples. In this visualization, circles denote the embedded space of samples, while stars represent class-level prototypes. Notably, we observe that few samples of the new class are closely clustered together. This is due to the SGN refining features through inter-sample associations. Furthermore, the CGN aids in aligning categories with strong similarities, fostering connections between old and new classes. The visualization reinforces that class-level attributes of both old and new classes remain distinguishable.
2) Fig.~ref{fig:tsne}(b): This panel shows the application of the trained FSCIL task to the test set. Notably, the use of S2C enhances prototype adaptation and fine-tunes the decision boundary between old and new classes.
Overall, these visualizations underscore the efficacy of the S2C approach in adapting prototypes and refining decision boundaries for effective FSCIL tasks on the CUB-200-2011 dataset.

\section{Conclusion}
In this paper, we studied the FSCIL problems from the perspective of building the relationships between sample-level to class-level graph. 
We proposed a novel Sample-to-Class Graph Network (S2C) which consists of Sample-level Graph Network (SGN) and Class-level Graph Network (CGN). SGN is used to build the relationship between samples in the N-way K-shot few-shot tasks to mine more favorable refined features. CGN is used to construct the context relationship between old and novel classes. 
Moreover, a S2C multi-stage training strategy was employed to improve the adaptation of S2C to novel classes. In general, S2C enhances the long-term learning ability of the deep learning model by simultaneously overcoming the catastrophic forgetting and generalization problems. 
Experimental result on benchmark datasets showed that our model is superior in both performance and adaptability than the state of the art methods. 
In our future work, we plan to enhance the edge information between graph nodes by incorporating additional data to further investigate the relationships and dependencies within few-shot data and construct multiple mapping relationships from sample-level graph to class-level graph to establish a more stable and robust multi-task relationship.

\bibliographystyle{IEEEtran}
\bibliography{icme2023template}

\begin{thebibliography}{10}
\providecommand{\url}[1]{#1}
\csname url@samestyle\endcsname
\providecommand{\newblock}{\relax}
\providecommand{\bibinfo}[2]{#2}
\providecommand{\BIBentrySTDinterwordspacing}{\spaceskip=0pt\relax}
\providecommand{\BIBentryALTinterwordstretchfactor}{4}
\providecommand{\BIBentryALTinterwordspacing}{\spaceskip=\fontdimen2\font plus
\BIBentryALTinterwordstretchfactor\fontdimen3\font minus \fontdimen4\font\relax}
\providecommand{\BIBforeignlanguage}[2]{{%
\expandafter\ifx\csname l@#1\endcsname\relax
\typeout{** WARNING: IEEEtran.bst: No hyphenation pattern has been}%
\typeout{** loaded for the language `#1'. Using the pattern for}%
\typeout{** the default language instead.}%
\else
\language=\csname l@#1\endcsname
\fi
#2}}
\providecommand{\BIBdecl}{\relax}
\BIBdecl

\bibitem{aljundi2018memory}
R.~Aljundi, F.~Babiloni, M.~Elhoseiny, M.~Rohrbach, and T.~Tuytelaars, ``Memory aware synapses: Learning what (not) to forget,'' in \emph{ECCV}, 2018.

\bibitem{castro2018end}
F.~M. Castro, M.~J. Mar{\'\i}n-Jim{\'e}nez, N.~Guil, C.~Schmid, and K.~Alahari, ``End-to-end incremental learning,'' in \emph{ECCV}, 2018.

\bibitem{chen2018closer}
W.-Y. Chen, Y.-C. Liu, Z.~Kira, Y.-C.~F. Wang, and J.-B. Huang, ``A closer look at few-shot classification,'' in \emph{ICLR}, 2018.

\bibitem{chen2021eckpn}
C.~Chen, X.~Yang, C.~Xu, X.~Huang, and Z.~Ma, ``Eckpn: Explicit class knowledge propagation network for transductive few-shot learning,'' in \emph{CVPR}, 2021.

\bibitem{cheraghian2021semantic}
A.~Cheraghian, S.~Rahman, P.~Fang, S.~K. Roy, L.~Petersson, and M.~Harandi, ``Semantic-aware knowledge distillation for few-shot class-incremental learning,'' in \emph{CVPR}, 2021.

\bibitem{du2022agcn}
K.~Du, F.~Lyu, F.~Hu, L.~Li, W.~Feng, F.~Xu, and Q.~Fu, ``Agcn: augmented graph convolutional network for lifelong multi-label image recognition,'' in \emph{ICME}.\hskip 1em plus 0.5em minus 0.4em\relax IEEE, 2022.

\bibitem{gomes2017survey}
H.~M. Gomes, J.~P. Barddal, F.~Enembreck, and A.~Bifet, ``A survey on ensemble learning for data stream classification,'' \emph{CSUR}, vol.~50, no.~2, 2017.

\bibitem{hou2019learning}
S.~Hou, X.~Pan, C.~C. Loy, Z.~Wang, and D.~Lin, ``Learning a unified classifier incrementally via rebalancing,'' in \emph{CVPR}, 2019.

\bibitem{joseph2021incremental}
K.~Joseph, J.~Rajasegaran, S.~Khan, F.~S. Khan, and V.~N. Balasubramanian, ``Incremental object detection via meta-learning,'' \emph{TPAMI}, 2021.

\bibitem{lee2019meta}
K.~Lee, S.~Maji, A.~Ravichandran, and S.~Soatto, ``Meta-learning with differentiable convex optimization,'' in \emph{CVPR}, 2019.

\bibitem{liu2020negative}
B.~Liu, Y.~Cao, Y.~Lin, Q.~Li, Z.~Zhang, M.~Long, and H.~Hu, ``Negative margin matters: Understanding margin in few-shot classification,'' in \emph{ECCV}, 2020.

\bibitem{ma2021transductive}
Y.~Ma, S.~Bai, S.~An, W.~Liu, A.~Liu, X.~Zhen, and X.~Liu, ``Transductive relation-propagation network for few-shot learning,'' in \emph{IJCAI}, 2021.

\bibitem{lyu2021multi}
F.~Lyu, S.~Wang, W.~Feng, Z.~Ye, F.~Hu, and S.~Wang, ``Multi-domain multi-task rehearsal for lifelong learning,'' in \emph{AAAI}, vol.~35, no.~10, 2021.

\bibitem{pernici2021class}
F.~Pernici, M.~Bruni, C.~Baecchi, F.~Turchini, and A.~Del~Bimbo, ``Class-incremental learning with pre-allocated fixed classifiers,'' in \emph{ICPR}.\hskip 1em plus 0.5em minus 0.4em\relax IEEE, 2021.

\bibitem{pham2021dualnet}
Q.~Pham, C.~Liu, and S.~Hoi, ``Dualnet: Continual learning, fast and slow,'' \emph{NIPS}, 2021.

\bibitem{rebuffi2017icarl}
S.-A. Rebuffi, A.~Kolesnikov, G.~Sperl, and C.~H. Lampert, ``icarl: Incremental classifier and representation learning,'' in \emph{CVPR}, 2017.

\bibitem{rusu2018meta}
A.~A. Rusu, D.~Rao, J.~Sygnowski, O.~Vinyals, R.~Pascanu, S.~Osindero, and R.~Hadsell, ``Meta-learning with latent embedding optimization,'' in \emph{ICLR}, 2018.

\bibitem{shi2021overcoming}
G.~Shi, J.~Chen, W.~Zhang, L.-M. Zhan, and X.-M. Wu, ``Overcoming catastrophic forgetting in incremental few-shot learning by finding flat minima,'' \emph{NIPS}, 2021.

\bibitem{sun2022exploring}
Q.~Sun, F.~Lyu, F.~Shang, W.~Feng, and L.~Wan, ``Exploring example influence in continual learning,'' 2022.

\bibitem{tao2020few}
X.~Tao, X.~Hong, X.~Chang, S.~Dong, X.~Wei, and Y.~Gong, ``Few-shot class-incremental learning,'' in \emph{CVPR}, 2020.

\bibitem{vaswani2017attention}
A.~Vaswani, N.~Shazeer, N.~Parmar, J.~Uszkoreit, L.~Jones, A.~N. Gomez, {\L}.~Kaiser, and I.~Polosukhin, ``Attention is all you need,'' \emph{NIPS}, 2017.

\bibitem{verma2019manifold}
V.~Verma, A.~Lamb, C.~Beckham, A.~Najafi, I.~Mitliagkas, D.~Lopez-Paz, and Y.~Bengio, ``Manifold mixup: Better representations by interpolating hidden states,'' in \emph{ICML}.\hskip 1em plus 0.5em minus 0.4em\relax PMLR, 2019.

\bibitem{vinyals2016matching}
O.~Vinyals, C.~Blundell, T.~Lillicrap, D.~Wierstra \emph{et~al.}, ``Matching networks for one shot learning,'' \emph{NIPS}, 2016.

\bibitem{wang2022learning}
Z.~Wang, Z.~Zhang, C.-Y. Lee, H.~Zhang, R.~Sun, X.~Ren, G.~Su, V.~Perot, J.~Dy, and T.~Pfister, ``Learning to prompt for continual learning,'' in \emph{CVPR}, 2022.

\bibitem{yang2020dpgn}
L.~Yang, L.~Li, Z.~Zhang, X.~Zhou, E.~Zhou, and Y.~Liu, ``Dpgn: Distribution propagation graph network for few-shot learning,'' in \emph{CVPR}, June 2020.

\bibitem{zhang2020deepemd}
C.~Zhang, Y.~Cai, G.~Lin, and C.~Shen, ``Deepemd: Few-shot image classification with differentiable earth mover's distance and structured classifiers,'' in \emph{CVPR}, 2020.

\bibitem{zhang2021few}
C.~Zhang, N.~Song, G.~Lin, Y.~Zheng, P.~Pan, and Y.~Xu, ``Few-shot incremental learning with continually evolved classifiers,'' in \emph{CVPR}, 2021.

\bibitem{zhao2020maintaining}
B.~Zhao, X.~Xiao, G.~Gan, B.~Zhang, and S.-T. Xia, ``Maintaining discrimination and fairness in class incremental learning,'' in \emph{CVPR}, 2020.

\bibitem{zhao2022deep}
T.~Zhao, Z.~Wang, A.~Masoomi, and J.~Dy, ``Deep bayesian unsupervised lifelong learning,'' \emph{Neural Networks}, 2022.

\bibitem{zhou2021co}
D.-W. Zhou, H.-J. Ye, and D.-C. Zhan, ``Co-transport for class-incremental learning,'' in \emph{Proceedings of the 29th ACM International Conference on Multimedia}, 2021.

\bibitem{zhou2021learning}
D.-W. Zhou, Y.~Yang, and D.-C. Zhan, ``Learning to classify with incremental new class,'' \emph{TNNLS}, 2021.

\bibitem{zhou2022few}
D.-W. Zhou, H.-J. Ye, L.~Ma, D.~Xie, S.~Pu, and D.-C. Zhan, ``Few-shot class-incremental learning by sampling multi-phase tasks,'' \emph{TPAMI}, 2022.

\bibitem{zhou2022forward}
D.-W. Zhou, F.-Y. Wang, H.-J. Ye, L.~Ma, S.~Pu, and D.-C. Zhan, ``Forward compatible few-shot class-incremental learning,'' in \emph{CVPR}, 2022.

\bibitem{zhu2021prototype}
F.~Zhu, X.-Y. Zhang, C.~Wang, F.~Yin, and C.-L. Liu, ``Prototype augmentation and self-supervision for incremental learning,'' in \emph{CVPR}, 2021.

\bibitem{zhu2021self}
K.~Zhu, Y.~Cao, W.~Zhai, J.~Cheng, and Z.-J. Zha, ``Self-promoted prototype refinement for few-shot class-incremental learning,'' in \emph{CVPR}, 2021.

\bibitem{Ji2023MCNet}
Z.~Ji, Z.~Hou, X.~Liu, Y.~Pang, and X.~Li, ``Memorizing complementation network for few-shot class-incremental learning,'' \emph{TIP}, 2023.

\bibitem{Xu2023MFS3}
X.~Xu, S.~Niu, Z.~Wang, W.~Guo, L.~Jing, and H.~Yang, ``Multi-feature space similarity supplement for few-shot class incremental learning,'' \emph{Knowledge-Based Systems}, 2023.

\bibitem{BTVR}
A.~Srinivas, T.-Y. Lin, N.~Parmar, J.~Shlens, P.~Abbeel, and A.~Vaswani, ``Bottleneck transformers for visual recognition,'' in \emph{CVPR}, 2021.

\bibitem{MetaSearch}
Q.~Wang, X.~Liu, W.~Liu, A.-A. Liu, W.~Liu, and T.~Mei, ``Metasearch: Incremental product search via deep meta-learning,'' \emph{TIP}, 2020.

\bibitem{Afn}
K.~Li, Y.~Zhang, K.~Li, and Y.~Fu, ``Adversarial feature hallucination networks for few-shot learning,'' in \emph{CVPR}, 2020.

\bibitem{ISM}
z.~Ji, Z.~Hou, X.~Liu, Y.~Pang, and J.~Han, ``Information symmetry matters: A modal-alternating propagation network for fewshot learning,'' \emph{TIP}, 2022.

\bibitem{MgSvF}
H.~Zhao, Y.~Fu, M.~Kang, Q.~Tian, F.~Wu, and X.~Li, ``Mgsvf: Multigrained slow vs. fast framework for few-shot class-incremental learning,'' \emph{TIPAMI}, 2020.

\bibitem{pretrain}
A.~Chowdhury, J.~Mingchao, S.~Chaudhuri, and C.~Jermaine, ``"few-shot image classification: Just use a library of pre-trained feature extractors and a simple classifier,'' in \emph{ICCV}, 2021.

\bibitem{SDC}
L.~Yu, T.~Twardowski, X.~Liu, l.~Herranz, K.~Wang, Y.~Cheng, and S.~Jui, ``Semantic drift compensation for class-incremental learning,'' in \emph{CVPR}, 2020.

\bibitem{Topology}
T.~Xiaoyu, C.~Xinyuan, H.~Xiaopeng, W.~Xing, and G.~Yihong, ``Topology-preserving class-incremental learning,'' in \emph{ECCV}, 2020.

\bibitem{CILKD}
K.~Minsoo, P.~Jaeyoo, and H.~Bohyung, ``Class-incremental learning by knowledge distillation with adaptive feature consolidation,'' in \emph{CVPR}, 2022.

\bibitem{EGNN}
K.~Jongmin, K.~Taesup, K.~Sungwoong, and C.~D. Yoo, ``Edge-labeling graph neural network for few-shot learning,'' in \emph{CVPR}, 2019.

\bibitem{GNN}
F.~Scarselli, M.~Gori, A.~C. Tsoi, M.~Hagenbuchner, and G.~Monfardini, ``The graph neural network model,'' \emph{IEEE Transactions on Neural Networks}, 2009.

\bibitem{Metalearning}
C.~Yinbo, L.~Zhuang, X.~Huijuan, D.~Trevor, and W.~Xiaolong, ``Meta-baseline: Exploring simple meta-learning for few-shot learning,'' in \emph{ICCV}, 2021.

\bibitem{MetaFSCIL}
C.~Zhixiang, G.~Li, L.~Huan, W.~Yang, Y.~Yuanhao, and J.~Tang, ``Metafscil: A meta-learning approach for few-shot class incremental learning,'' in \emph{CVPR}, 2022.

\bibitem{CalibratingCNN}
Y.~Peng, R.~Shaogang, Z.~Yang, and L.~Ping, ``Calibrating cnns for few-shot meta learning,'' in \emph{WACV}, 2022.

\bibitem{cifar}
K.~Alex and H.~Geoffrey, ``Learning multiple layers of features from tiny images,'' in \emph{CVPR}, 2009.

\bibitem{MiniImagenet}
R.~Olga, D.~Jia, S.~Hao, K.~Jonathan, S.~Sanjeev, M.~Sean, H.~Zhiheng, K.~Andrej, K.~Aditya, and B.~Michael, ``Imagenet large scale visual recognition challenge,'' in \emph{IJCV}, 2015.

\bibitem{Cub}
W.~Catherine, B.~Steve, W.~Peter, P.~Pietro, and B.~Serge, ``The caltech-ucsd birds-200-2011 dataset,'' in \emph{ECCV}, 2011.

\end{thebibliography}










\appendix
\begin{figure*}[ht]
	\centering
	\includegraphics[width=1.\linewidth]{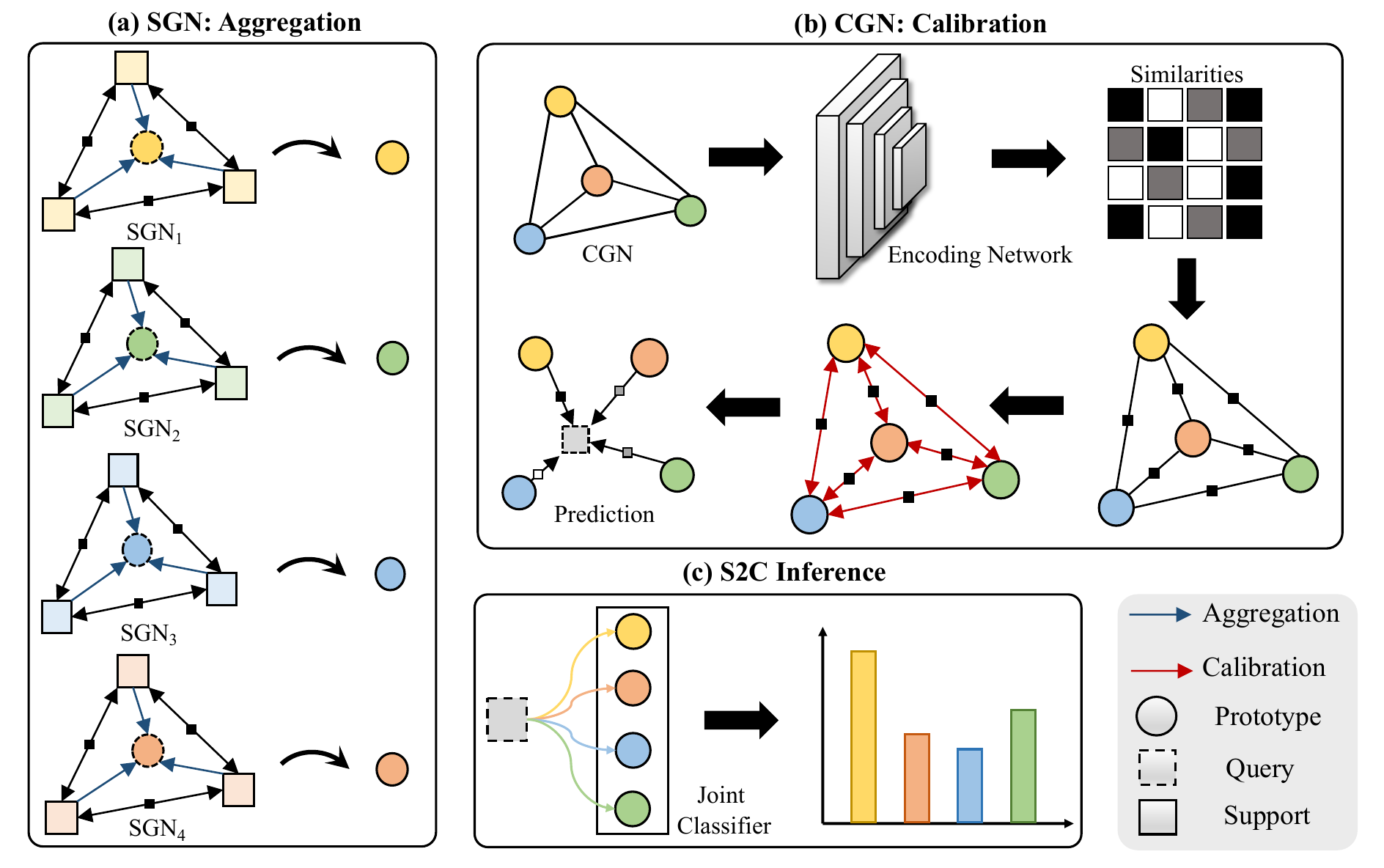}
	\caption{Illustration of S2C.  }
	\label{fig:Visualization}
\end{figure*}
\section{Introduction about Compared Methods}
In this section, we provide a comprehensive introduction to the comparative methodologies employed in the primary paper. These methodologies are enumerated as follows:
\begin{itemize}
    \item \textbf{Finetune.} When confronted with the challenge of a few-shot incremental session, it employs a straightforward approach of optimizing the cross-entropy across these limited few-shot instances. However, this approach is prone to experiencing catastrophic forgetting.
    \item \textbf{iCaRL.} During the training of a novel incremental task, it integrates the cross-entropy loss with the knowledge distillation loss. This inclusion of knowledge distillation serves to enable the model to sustain its discriminative capacity over previously acquired knowledge.
    \item \textbf{EEIL.} Incorporating an additional balanced fine-tuning procedure beyond iCaRL, this approach employs a balanced dataset to refine the model, effectively mitigating bias.
    \item \textbf{Rebalancing}. Implements cosine normalization, feature-wise knowledge distillation, and contrastive learning techniques to enhance the model's capabilities and mitigate the risk of catastrophic forgetting.
    \item \textbf{TOPIC.} Customizes the few-shot class-incremental learning task utilizing a neural gas network, which preserves the topological structure of the feature manifold across distinct classes.
    \item \textbf{Decoupled-Cosine}. Much like the DecoupledDeepEMD approach, it disentangles the training procedures for embedding and classification. Additionally, it employs cosine distance calculation during the inference phase.
    \item \textbf{Decoupled-DeepEMD.} Similar to the DecoupledCosine method, this approach separates the training processes for embedding and classification. Notably, it differentiates from DecoupledCosine by incorporating a negative margin softmax function during model pretraining. Furthermore, it employs cosine distance calculation during the inference phase.
    \item \textbf{F2M.} Search for flat local minima of the base training objective function and then fine-tune the model parameters within the flat region on new tasks to overcome catastrophic forgetting
    \item \textbf{CEC.} During the base session, an additional graph model is trained using pseudo-incremental learning sampling. This graph model is designed to adapt the embeddings of both established class prototypes and newly introduced class prototypes. This adaptability is transferable and extends to the incremental learning process.
    \item \textbf{LIMIT.} Simulate fake FSCIL tasks and prepare the model for future FSCIL tasks. encode the inductive bias into the meta-calibration module, which helps to calibrate between classifiers and the few-shot prototypes.
    \item \textbf{FACT.} Suggest learning prospectively to prepare for future updates and  preassign virtual prototypes in the embedding space to reserve the space for new classes from two aspects.
    \item \textbf{MCNet.} Ensemble multiple embedding networks to realize the complementation with each other. With the Structure-Wise Complementation (SWC) and the Task-Wise Complementation (TWC), the whole model with different embedding networks memorizes more complete knowledge when updated in incremental sessions.
    \item \textbf{MFS3.} A multi-feature space similarity supplement (MFS3) is introduced to alleviate the problems of forgetting and adaption. MFS3 includes inter-feature space similarity supplement (IFS3) which focus on boundary-sensitive sample points and outer-feature space similarity supplement (OFS3) which can utilize the supplement of base feature space with new feature space to rejudge the sample points. 
\end{itemize}

\begin{table*}[ht]
\scriptsize
    \centering
    \caption{Ablation results of S2C on CIFAR100 dataset.}\label{lab:ab_cifar}
    \resizebox{1.\linewidth}{!}{
        \begin{tabular}{l cccccccccc cc  }
            \bottomrule
            \multirow{2}{*}{\textbf{Components}} &\multicolumn{9}{c}{\textbf{Accuracy in each session (\%)}}  & \multirow{2}{*}{\textbf{PD}} & \multirow{2}{*}{\textbf{Improvement}}\\
            \cline{2-10} 
            &\textbf{0} &\textbf{1}  &\textbf{2}&\textbf{3} &\textbf{4}&\textbf{5}  &\textbf{6}&\textbf{7}&\textbf{8} & &\\ 
            \hline
            baseline           &  69.75  &  65.06  &  61.20  &   57.21  &   53.88  &   51.40  &   48.80  &   46.84  &   44.41  &  25.34  &  \textbf{+3.38}  \\
            baseline+SGN           &  73.07  &  68.88  &  65.26  &  61.19  &  58.09  &  55.57  &  53.22  &  51.34  &  49.14  &  23.93  &  \textbf{+1.97}  \\
            baseline+CGN            &  73.81  &  71.09  &  66.87  &  62.89  &  59.70  &  56.77  &  54.67  &  52.52  &  50.23  &  23.58  & \textbf{+1.62}   \\
            \hline
            baseline+SGN+CGN       &  \textbf{75.15}  &  \textbf{73.07}  &  \textbf{68.31}  &  \textbf{64.61}  &  \textbf{61.94}  &  \textbf{59.41}  &  \textbf{57.62}  &  \textbf{55.62}  &  \textbf{53.19}  &  \textbf{21.96}    \\
            \toprule
        \end{tabular}}
\end{table*}

\begin{table*}[ht]
\scriptsize
    \centering
    \caption{Ablation results of S2C on miniImageNet dataset.}\label{lab:ab_mini}
    \resizebox{1.\linewidth}{!}{
        \begin{tabular}{l cccccccccc cc  }
            \bottomrule
            \multirow{2}{*}{\textbf{Components}} &\multicolumn{9}{c}{\textbf{Accuracy in each session (\%)}}  & \multirow{2}{*}{\textbf{PD}} & \multirow{2}{*}{\textbf{Improvement}}\\
            \cline{2-10} 
            &\textbf{0} &\textbf{1}  &\textbf{2}&\textbf{3} &\textbf{4}&\textbf{5}  &\textbf{6}&\textbf{7}&\textbf{8} & &\\ 
            \hline
            baseline           &  69.17  &  63.74  &  59.13  &   55.13  &   51.63  &   48.44  &   45.67  &   43.58  &   41.74  &  27.43  &  \textbf{+6.18}  \\
			baseline+SGN           &  71.68  &  68.95  &  65.67  &  62.36  &  59.08  &  55.8  &  52.98  &  51.18  &  49.42  &  22.26  &  \textbf{+1.01}  \\
			baseline+CGN            &  71.5  &  67.75  &  63.64  &  60.43  &  57.0  &  53.85  &  51.11  &  49.19  &  47.81  &  23.69  & \textbf{+2.44}   \\
	
			\hline
			baseline+SGN+CGN       	   &  \textbf{73.25}  &  \textbf{71.57}  &  \textbf{67.46}  &  \textbf{64.01}  &  \textbf{61.04}  &  \textbf{58.41}  &  \textbf{55.62}  &  \textbf{53.62}  &  \textbf{52.00}  &  \textbf{21.25}     \\
            \toprule
        \end{tabular}}
\end{table*}

\begin{table*}[ht]
\scriptsize
\centering
\caption{Ablation results of S2C on CUB200 dataset.}\label{lab:ab_cub}
\resizebox{1\linewidth}{!}{
 \begin{tabular}{l cccccccccccc cc  }
  \toprule
  \multirow{2}{*}{\textbf{Method}} &\multicolumn{11}{c}{\textbf{Accuracy in each session(\%)}}  & \multirow{2}{*}{\textbf{PD}} & \multirow{2}{*}{\textbf{Improve}}\\
  \cline{2-12}
  & \textbf{0} &\textbf{1}  &\textbf{2}&\textbf{3} &\textbf{4}&\textbf{5}  &\textbf{6}&\textbf{7}&\textbf{8}&\textbf{9}&\textbf{10}  & & \\ 
  \hline
  baseline           &  68.68  &  63.85  &  59.43  &54.68  &  51.19  &  47.88  &  45.65  &  44.07  &  42.17  &   40.63  &   39.33    &  29.35  &  \textbf{+10.89}  \\
  baseline+SGN              &  74.35  &  70.69  &  66.68  &  62.34  &  59.76  &  56.54  &  54.61  &  52.52  &  50.73  &  49.20  &  47.60  &  26.75  &  \textbf{+8.29}  \\
  baseline+CGN              &  74.85  &  71.94  &  68.50  &  65.50  &  62.43  &  59.27  &  57.73  &  55.81  &  54.83  &  53.52  &  52.28  &  22.57  &  \textbf{+4.11}  \\
  \hline
  baseline+SGN+CGN   &  \textbf{75.92}  &  \textbf{73.57}  &  \textbf{71.67}  &  \textbf{68.01}  &  \textbf{66.94}  &  \textbf{63.61}  &  \textbf{62.22}  &  \textbf{61.42}  &  \textbf{59.79}   & \textbf{58.56} & \textbf{57.46}  &  \textbf{18.46}  \\
  \toprule
	\end{tabular}}
\end{table*}

It's important to observe that Finetune, Pre-Allocated RPC, iCaRL, EEIL, and Rebalancing are conventional Class-Incremental Learning (CIL) methods. Our empirical experiments, as discussed in the primary paper, demonstrate that these traditional methods are not well-suited for FSCIL. Among other state-of-the-art methods in FSCIL, \ie, LIMIT, FACT, MCNet, MFS3, our proposed S2C consistently outperforms them across a range of performance metrics.

\section{ Detailed results about ablation studies}
In the primary paper, we conducted a comprehensive analysis of the significance of each constituent within the S2C approach across the Mini-ImageNet, CIFAR-100, and CUB-200-2011 datasets. The detailed outcomes on these benchmark datasets are presented in Table ~\ref{lab:ab_cifar}, ~\ref{lab:ab_mini}, ~\ref{lab:ab_cub}. 
The incorporation of the SGN module enhances the learning performance of FSL tasks. This enhancement isn't confined to FSL tasks but extends to the overall performance across various sessions, underscoring the importance of the SGN in FSL task learning. The baseline+SGN method has notably improved the learning effect of a single task compared to the baseline, thanks to SGN's powerful ability to build relationships for few-shot data.
Besides, the combination of the CGN module effectively addresses the issue of catastrophic forgetting observed in the baseline model during FSCIL tasks. In the CIFAR100 and CUB200 datasets, our baseline+CGN method consistently outperforms the baseline+SGN method in each session, contributing significantly to the reduction of the PD of the model. Leveraging both SGN and CGN, our S2C training strategy deploys sample-level graph and class-level graph in FSCIL scenario to address both the challenges of mitigating forgetting and alleviating overfitting. By integrating both the SGN and CGN modules, the approach not only improves the performance of FSL tasks but also considers long-dependence relationship that may arise due to data imbalance and other factors between old and new classes. Our inference is that both the SGN and CGN modules notably play pivotal roles in achieving success in FSCIL tasks. These findings align with the conclusions drawn in the main paper, thereby affirming the superior performance of S2C. 

\section{Illustration of the SGN and CGN}
In this section, we supplement the information on the detailed implementation of S2C in Fig.~\ref{fig:Visualization}, mainly including the construction details of SGN and CGN. 

As depicted in Fig.~\ref{fig:Visualization}(a), we incorporate the SGN to facilitate FSL task learning. SGN is responsible for aggregating samples belonging to the same class while distinguishing samples from different classes through the exploration of inter-sample relationships. This process aids in the extraction of refined class-level features. The sample representations are iteratively updated based on the relationship parameters between samples. Subsequently, we compute the embeddings and take the average for each class, resulting in the generation of class-level features. 

CGN is designed to capture contextual relationships between old and new classe- level features, allowing for adjustments to the embedding space of prototype features for new classes within the class-level graph. As shown in ~\ref{fig:Visualization}(b), in our implementation, we employ a combination of the Transformer and GNN. More precisely, we utilize the multi-head attention mechanism to establish connections between old and new classes and utilize the GNN to aggregate this information for the calibration of prototype features for new classes. The CGN constructs the edge relationships by calculating the similarity matrix between nodes, assigns corresponding weights, and then iteratively updates the nodes using multi-head self-attention mechanism. This process ultimately leads to making predictions for the Query samples based on the learned relationships and weights within the graph. 

After the calibration of all class level features by CGN, we can obtain a joint classifier and each class level feature will be used as a weight for this joint classifier to make the final prediction for Query samples based on the metric-based approach.

\end{document}